\documentclass[letterpaper]{article} 
\usepackage{aaai25}  
\usepackage{times}  
\usepackage{helvet}  
\usepackage{courier}  
\usepackage[hyphens]{url}  
\usepackage{graphicx} 
\usepackage{natbib}  
\usepackage{caption} 
\usepackage{algorithm}
\usepackage{algpseudocode}

\usepackage{graphicx}
\usepackage{array}
\usepackage{booktabs}
\usepackage{tabularx}
\usepackage{multirow}
\usepackage{verbatim}
\usepackage{colortbl}
\usepackage{arydshln}
\usepackage{xcolor}
\usepackage{amsmath}  
\usepackage{amssymb}  

\newcommand{\xia}[1]{{\color{black} #1}}  

\newcommand{\zhou}[1]{{\color{black}{#1}}} 
\newcommand{\zhourev}[1]{{\color{black}{#1}}} 

\usepackage{newfloat}
\usepackage{listings}
\DeclareCaptionStyle{ruled}{labelfont=normalfont,labelsep=colon,strut=off} 
\floatstyle{ruled}
\newfloat{listing}{tb}{lst}{}
\floatname{listing}{Listing}
\title{Improving Retrieval Augmented Language Model with Self-Reasoning}

\author{Yuan Xia$^{1}$, Jingbo Zhou$^{2,}$\thanks{Jingbo Zhou is the corresponding author.}, Zhenhui Shi$^{1}$, Jun Chen$^{1}$, Haifeng Huang$^{1}$}
 \affiliations{
     $^{1}$Baidu Inc., China $^{2}$Baidu Research, China\\
    \{xiayuan,zhoujingbo,shizhenhui,chenjun22,huanghaifeng\}@baidu.com
}
\usepackage{bibentry}

\begin{document}

\maketitle

\begin{abstract}
The Retrieval-Augmented Language Model (RALM) has demonstrated remarkable performance on knowledge-intensive tasks by integrating external knowledge during inference, which mitigates the factual hallucinations inherited in large language models (LLMs). Despite these advancements, challenges persist in the implementation of RALMs, particularly in terms of reliability and traceability. Specifically, the irrelevant document retrieval may result in unhelpful responses or even deteriorate the performance of LLMs, while the lack of appropriate citations in outputs complicates efforts to verify the trustworthiness of the models. To this end, we propose a novel self-reasoning framework aimed at improving the reliability and traceability of RALMs, whose core idea is to leverage reasoning trajectories generated by the LLM itself. The framework involves constructing self-reasoning trajectories through three processes: a relevance-aware process, an evidence-aware selective process, and a trajectory analysis process. We evaluated our framework across four public datasets (two short-form QA datasets, one long-form QA dataset, and one fact verification dataset) to demonstrate its superiority. Our method can outperform existing state-of-the-art models and achieve performance comparable with GPT-4, using only 2,000 training samples. 
\end{abstract}



\section{Introduction}

\begin{figure}[t]
\center
\includegraphics[width=0.48\textwidth]{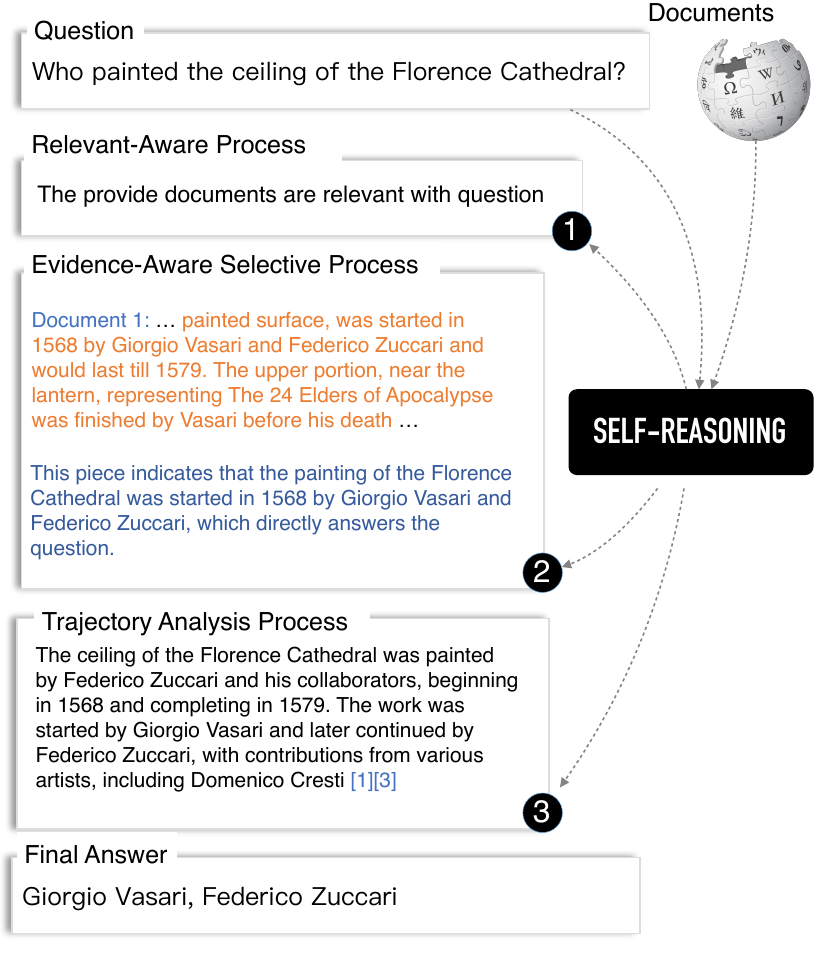}
\caption{An example of how \textsc{self-reasoning} framework generates reasoning trajectories. }
\label{fig:demo}
\end{figure}

The Retrieval-Augmented Language Model (RALM), also known as Retrieval-Augmented Generation (RAG), has become a crucial enhancement for Large Language Models (LLMs) by integrating external knowledge during inference.
Despite their \zhourev{advanced} capabilities in language understanding and generation \cite{brown2020language,touvron2023llama}, LLMs are prone to producing hallucinated and inaccurate content, especially in knowledge-intensive tasks \cite{ji2023survey}.
Augmenting LLMs with relevant information obtained from external sources like Wikipedia and search engines has proven effective in reducing these inaccuracies \cite{10.5555/3524938.3525306,lewis2020retrieval,borgeaud2022improving,izacard2022few,asai2024selfrag}.
This approach has proven effective in mitigating the factual hallucinations that are inherent in LLMs \cite{kwiatkowski-etal-2019-natural, petroni-etal-2021-kilt, ram-etal-2023-context}.

Nevertheless, there are still limitations associated with RALMs, particularly concerning reliability and traceability. Firstly, the reliability of the retrieved information remains a substantial concern.  Previous studies have shown that noisy retrieval can adversely affect the performance of an LLM \cite{menick2022teaching,li-etal-2023-large}, as irrelevant data can lead to misguided responses and disturb the model's ability to leverage its intrinsic knowledge effectively. Secondly, the interpretability and traceability of outputs generated by RALMs need to be improved. Although RALMs incorporate retrieved documents during both the training and inference phases, they may fail to explicitly cite these documents, thus complicating the process of tracing and verifying the claims made by LLMs. To improve the retrieval robustness, recent studies have explored incorporating external tools such as natural language inference (NLI) models \cite{honovich2022true} and document summarization models during inference \cite{yoran2023making,xu2024recomp}. However, the effectiveness of these external tools largely influences the overall performance of RALMs. Additionally, training and optimizing these auxiliary models require additional costs. Consequently, identifying the most appropriate training and selection methods for NLI and summarization models remains a critical challenge in leveraging these approaches.

\begin{figure*}
    \center
    \includegraphics[width=0.95\linewidth]{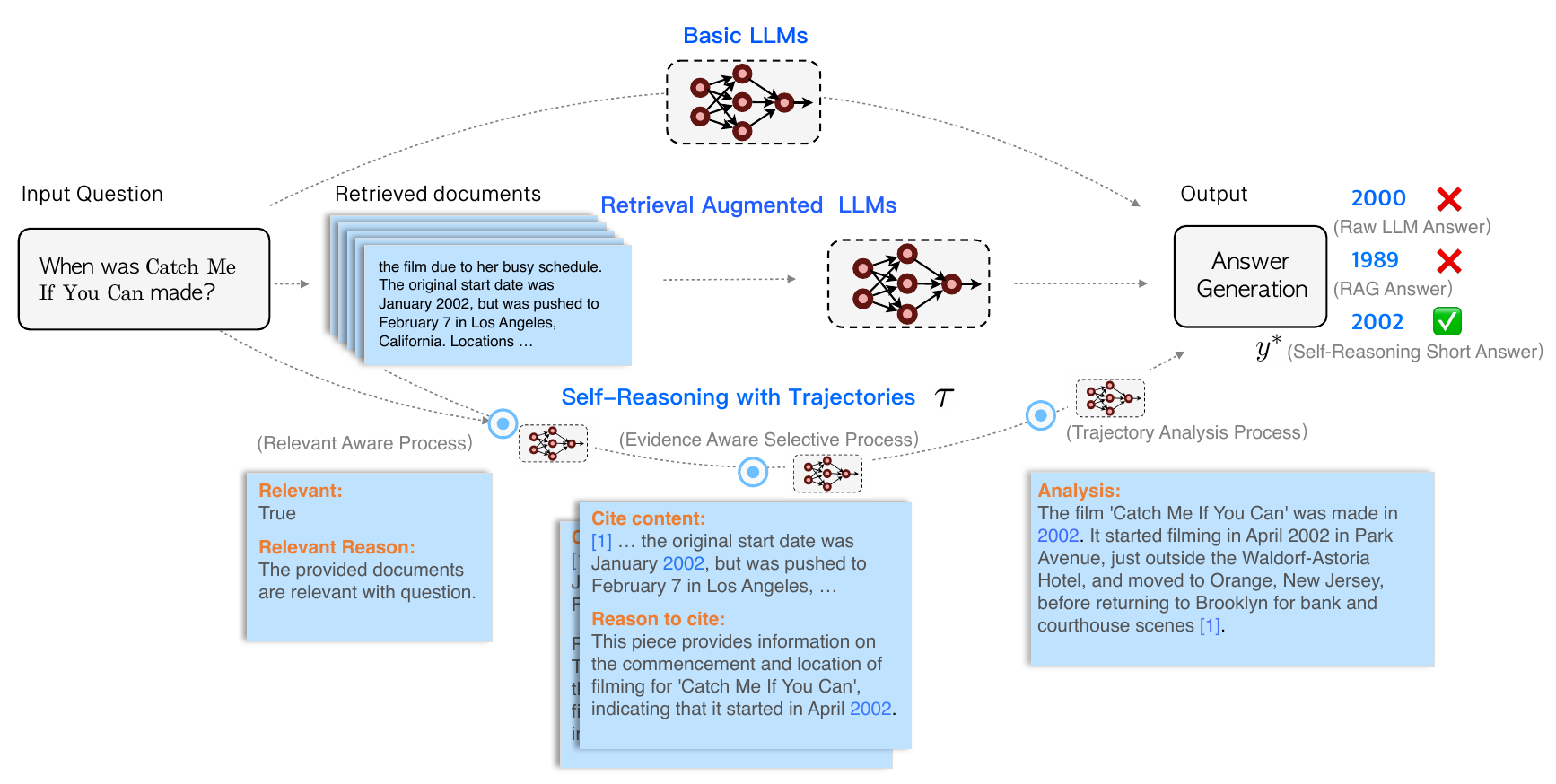}
    \caption{An illustration of the \textsc{self-reasoning} framework. The upper is the basic LLMs which answer the question by inherent knowledge. The middle is the standard retrieval augmented LMs, which use retrieved documents to help answer the question. The bottom is our \textsc{self-reasoning} framework which uses self-generated reason trajectories to output answers.}
    \label{fig:arch}
\end{figure*}

To address the above limitations, we propose a novel end-to-end \textsc{self-reasoning} framework to improve the performance of RALMs. For convenience, we will also refer to this framework as \textsc{self-reasoning} RAG and use the terms interchangeably. 
Our intuition is that the explicit self-reasoning trajectory crafted by LLMs can improve both the retrieval robustness and accuracy in question answering. 
During the pre-training phase, while an LLM primarily focuses on knowledge acquisition, it does not learn to reason from retrieved documents to generate answers. To address this, a feasible approach is to incorporate reasoning trajectories into a post-training phase. Such an approach could potentially teach the model to reason and distinguish relevant and irrelevant documents, thereby enhancing its query response accuracy.
An example of how our \textsc{self-reasoning} framework generates reasoning trajectories is illustrated in Figure \ref{fig:demo}.  In contrast, as shown in the middle part of Figure \ref{fig:arch}, the conventional RALM methods gather all documents in a non-selective manner, leading to the distraction of the LLM by irrelevant content and consequently resulting in the generation of erroneous answers.

Our framework constructs self-reasoning trajectories comprising three processes: 1) a \textit{Relevance-Aware Process} (RAP), which instructs the LLM to judge the relevance between the retrieved documents and the question, 2) an \textit{Evidence-Aware Selective Process} (EAP), which directs the LLM to choose \zhou{and} cite relevant documents, and \zhou{then} automatically select snippets of key sentences as \textit{evidence} from the cited documents, 3) a \textit{Trajectory Analysis Process} (TAP), which requires the LLM to synthesize a concise analysis based on all gathered self-reasoning trajectories generated by previous two processes and subsequently provide the final inferred answer. Furthermore, we propose a gradual training method by employing stage-wise masking strategies to enhance the performance of our framework.
We summarize our contributions as follows:
\begin{itemize}

\item We propose a novel end-to-end \textsc{self-reasoning} framework that improves the robustness of RALMs by leveraging reasoning trajectories generated by the LLM itself, without the need for external tools.

\item We carefully design three processes to enhance the interpretability and traceability of RALMs by requiring LLMs to explicitly generate snippets and citations from documents, and further explain the reason why cited documents can help answer the question. 

\item We evaluate our framework on four public datasets (two short-form QA, one long-form QA, and one fact verification),
demonstrating that our method surpasses existing state-of-the-art models in performance using only 2,000 training samples.

\end{itemize}

\section{Related Work}

\subsection{Retrieval-augmented LMs}
Many studies have investigated augmenting the performance of LLMs with externally retrieved information \cite{izacard2022few,10.5555/3524938.3525306,borgeaud2022improving} and some of them pre-train language models with retrieved passages. 
For works focusing on RALMs with citations, \citet{menick2022teaching,nakano2021webgpt} instruct or train an LLM to answer questions with retrieved documents while providing citations. 
\citet{gao-etal-2023-enabling} proposes an end-to-end system to retrieve supporting evidence and generate answers with citations, while only focusing on prompting without updating their model weights. Other works instruct or fine-tune LLMs to use external tools to retrieve dynamically \cite{schick2023toolformer, yao2023react, jiang-etal-2023-active}, which offers an adaptive method of when and what to search. \citet{gao-etal-2023-rarr} improves the attribution and factuality of language models by taking outputs of LLMs and applying a post-process retrieve-and-edit approach. 

\subsection{Robustness for RALMs}

To improve the robustness of RALMs, previous works can be divided into two categories. The first category utilizes retrieved documents to enhance the Chain of Thought (CoT). For example, IRCoT \cite{trivedi2023interleaving} iteratively uses retrieved documents to generate CoT, which is then used to retrieve further documents in subsequent steps. ReAct \cite{yao2023react} introduces an iterative CoT paradigm that integrates reasoning with search results. However, irrelevant retrievals may produce misguided CoT, adversely affecting LLM performance \cite{menick2022teaching,li-etal-2023-large}.

To address the issue of irrelevant retrieval information, the second category proposes using external modules to process retrieved documents during inference. For instance, \citet{yoran2023making} utilize a natural language inference model to filter out irrelevant documents, \citet{yan2024corrective} employ a retrieval evaluator to classify documents based on their quality, and \citet{xu2024recomp} and \citet{yu2023chain} apply models to filter out or compress retrieved documents. \citet{baek-etal-2023-knowledge-augmented-language} deploy a separate small language model as a verifier to detect and correct errors in LLMs during retrieval. A method presented by \citet{asai2024selfrag}, which appears most similar to our approach, develops a technique that instructs models to retrieve information using specifically designed reflection tokens. However, this approach needs to train extra critic models and generator models to predict the reflection tokens, which requires tens of thousands of extra training samples.

Unlike the second group of works, which rely on external tools or additional modules to eliminate irrelevant information, the \textsc{self-reasoning} RAG method integrates self-reasoning directly into the model’s architecture, thereby enhancing the performance of LLMs and providing a more efficient and scalable solution. Further related works on LLMs for reasoning are discussed in the Appendix.

\section{Preliminary}

We formally define the problem of retrieval augmented generation with self-reasoning. Given a query $q$ and a corpus of documents $\mathcal{D}$, an LLM-generated answer with $m$ statements and $n$ tokens can be defined as $y=(s_1, s_2, \cdots, s_m) = (w_1, w_2, \cdots, w_n)$, where $s_i$ is the $i$-th statement and $w_j$ is the $j$-th token in the generated answer. In addition, for long-form QA settings, each statement $s_i$ should cite a list of documents $C_i = \{ c^{(1)}_i, c^{(2)}_i, . . .\}$, where $c^{(k)}_i \in \mathcal{D}$. In our work, we train an LLM (e.g. LLaMA2) to first generate reasoning trajectories $\tau$ through self-reasoning and then to generate answers ${y}^{*}$ (short-form answers) on condition of $\tau$. The model output is $y = {\rm{concat}} (\tau, y^*)$, which is the concatenation of $\tau$ and $y^*$. Note that the generations of $\tau$ and $y^*$  are done in a single pass within the \textsc{self-reasoning} framework.

\section{Method} \label{method}

Here we provide a detailed implementation of the self-reasoning process which involves three processes: 1) a \textit{Relevance-Aware Process} (RAP),  2) an \textit{Evidence-Aware Selective Process} (EAP), and 3) a \textit{Trajectory Analysis Process} (TAP). An illustration of our \textsc{self-reasoning} framework is shown in Figure \ref{fig:arch}. Additionally, we outline the process of data generation and quality control, and present the specifics of model training.

\subsection{Relevance-Aware Process}{\label{sec:rap}}

In this work, we choose DPR \cite{karpukhin-etal-2020-dense} and Contriever \cite{izacard2021unsupervised} as default retrievers $R$ to recall the top-$k$ relevant documents. When presented with a question and a set of documents, people can determine whether the question is relevant to the retrieved documents. Therefore, we first instruct the model to judge the relevance between the retrieved documents $\mathcal{D}$ and the given question $q$. We further request the model to explicitly generate reasons explaining why given documents are identified as relevant. The output should include two fields as \textit{relevant} and \textit{relevant reason}, as depicted in Figure \ref{fig:arch}. If all of the retrieved documents are irrelevant, the model should provide an answer based on the internal knowledge acquired during its pre-training phase. We define the self-reasoning trajectories generated by RAP as $\tau_{r}$.

\subsection{Evidence-Aware Selective Process}{\label{sec:eap}}
When answering a question, people generally first identify the crucial sentences from the provided documents and then cite or highlight them as key points. This process of citing the document facilitates reading comprehension and can serve as a technique for combining multiple short answers to address various aspects. While people may carry out this selective process and citation instantaneously, LLMs need to formulate the self-reasoning trajectories explicitly.

In our work, we require the LLM to explicitly state the reason why the selected sentence is supportive and plausible in answering the question. We define the selected sentence as \textit{evidence} in our paper. Specifically, after retrieving the top-$k$ documents, the self-reasoning method for \textit{Evidence-Aware Selective Process} can be formulated as follows: First, we instruct the LLM to choose relevant documents and automatically select snippets of key sentences for the selected documents. Then, we request the LLM to output the reason why the selected snippets can answer the question. The intermediate output is a list containing multiple contents, each content should include two fields, as \textit{cite content} and \textit{reason for cite}, which is illustrated in Figure \ref{fig:arch}. We define the self-reasoning trajectories generated by EAP as $\tau_{e}$.

\subsection{Trajectory Analysis Process}{\label{sec:tap}}
Finally, we consolidate all the self-reasoning trajectories ($\tau_{r}$ and $\tau_{e}$) in the previous processes together to form a chain of reasoning snippets, thereby enhancing the overall performance of the retrieval augmentation generation. Specifically, we ask the LLM to analyze the reasoning trajectories within itself and ultimately to output a concise analysis and a short answer. We instruct the LLM to output content with two fields as \textit{analysis} and \textit{answer}, which is shown in Figure \ref{fig:arch}. We define the self-reasoning trajectories generated by TAP as $\tau_{a}$. In this work, the \textit{analysis} output is defined as a long-form answer, and the \textit{answer} output is defined as a short-form answer. In the experiment section, we further explored the performance of long-form and short-form QA settings.

\subsection{Data Generation and Quality Control}{\label{sec:qc}}
\paragraph{\textbf{Training Data Generation.}}
For the \textit{Relevance-Aware Process} data generation, as manually labeling the relevant and irrelevant documents is label-intensive, we request GPT-4 \cite{openai2023gpt} to generate answers as ground truth. Specifically, we instruct GPT-4 to generate labels regarding irrelevant fields, and further to output the reasons why the given documents cannot answer the question. We concatenate the given question and the retrieved documents as positive samples. For negative samples, we randomly select a different question from the training set and retrieve the top-$k$ documents related to it. These documents are then concatenated with the initial question to form negative samples. To avoid order bias in the training data, we shuffle the order of the documents.

For the EAP and TAP data generation, manually annotating the citation and writing the self-reasoning process for each question is not feasible in practice. Therefore, we follow a similar process to RAP, we first instruct GPT-4 to generate a snippet of selected documents and subsequently output the reasoning process as trajectories. The method for constructing the EAP training data is the same as RAP except that the instructions given to GPT-4 are different. The details of the instructions are shown in the Appendix.

\paragraph{\textbf{Data Quality Control.}}
For training data generation, correct and comprehensive reasoning trajectories are very important. When training an LLM, the quality of the training samples is more important than the quantity \cite{zhou2023lima}. As we cannot guarantee the correctness of self-reasoning trajectories and citations by GPT-4, we develop two efficient methods to control the quality of data generation: 1) The first method is to use the off-the-shelf tools $\footnote{Tools are available at
\url{https://github.com/princeton-nlp/ALCE/tree/main}}$ in \citet{gao-etal-2023-enabling} to automatically verify the performance of data generation for document citations. We calculate the citation precision and recall score for each training sample and filter out scores lower than our pre-defined thresholds $\delta_{p}$ and $\delta_{r}$, for citation precision and recall, respectively. 2) Second, though the validation of self-reasoning trajectories and citations generated by GPT-4 is challenging, verifying the correctness of the final answer is straightforward. Therefore, we filter out the trajectories that lead to the incorrect answers and only keep the correct ones. We totally generate 10,000 training samples by GPT-4, after the filtering strategy by quality control, we finally keep 2,000 training samples with high quality. More details and pseudo-codes can be found in the Appendix.

\subsection{Model Training}{\label{sec:train}}
We train the self-reasoning RAG model $\phi$ by our constructed corpus which is augmented with self-reasoning trajectories $\tau$ using the standard language modeling objective, maximizing likelihood:
\begin{equation}\label{eq:likehood}
\max _{\mathcal{\phi}} \mathbb{E}_{(q, \tau, y) \sim \mathcal{D}_{sr}} \log p_{\mathcal{\phi}}(y \mid \tau, q) p_{\mathcal{\phi}}({\tau \mid q})
\end{equation}
where $\tau = \tau_{r} \oplus \tau_{e} \oplus \tau_{a}$ are the self-reasoning trajectories, $\oplus$ is a concatenation operator, $\tau_{r}, \tau_{e}, \tau_{a}$ are trajectories generated by above three processes respectively. $q$ is the provided question, and $y$ is the model output, including the intermediate reason trajectories and the final answer. $\mathcal{D}_{sr}$ is the training corpus augmented with self-reasoning trajectories.

During training, we observed that it is more challenging to ensure the correctness of an LLM with 13B parameters when generating long reasoning trajectories than short ones. We hypothesize that an LLM's effective reasoning length is limited and exceeding this limit might lead to error accumulation during the inference stage. Therefore, we propose a gradual training method by employing stage-wise masking strategies to gradually learn to generate long trajectories.

Specifically, we propose a stage-wise training process while we train the LLM stage by stage. In the first stage, we mask the trajectories produced by the next two stages (EAP and TAP) and train the model with a learning rate $r_{a}$. Then in the second stage, we only mask the trajectories generated by TAP and train the model with a learning rate $r_{b}$. Finally, we concatenate the reasoning trajectories from all stages and put them into a self-reasoning LLM for end-to-end training with a learning rate $r_{c}$. Hyper-parameters for training are described in the Appendix.

\section{Experiments}{\label{sec:exp}}

\begin{table*}[htb]
\small
\center
\begin{tabularx}{0.85\textwidth}{lcccccc}
 \toprule 
 \multirow{2}{*}{ \textbf{Models} } & \multicolumn{1}{c}{ NaturalQuestion} & \multicolumn{1}{c}{PopQA} & \multicolumn{1}{c}{FEVER} & \multicolumn{3}{c}{ASQA } \\
  \cmidrule(r){2-2}  \cmidrule(r){3-3}  \cmidrule(r){4-4} \cmidrule(l){5-7} 
& (acc) & (acc) & (acc) & (em-recall) & (precision) & (recall) \\
\midrule
\multicolumn{7}{>{\columncolor[gray]{.95}}c}{\textbf{\textit{Baselines without retrieval}}} \\

 \midrule 
LLaMA2\textsubscript{{7B}}& 19.2 & 18.4 & 23.2 &  10.2 &  - & - \\
 LLaMA2\textsubscript{{13B}}& 24.0 & 22.6 & 25.3 & 15.3 &  - & -  \\
LLaMA2\textsubscript{{7B-chat}} & 20.2 & 21.5 & 26.5 & 16.3  & - & -  \\
 LLaMA2\textsubscript{{13B-chat}} & 23.2 & 25.9 & 28.4 & 18.3  & - & - \\
  \midrule
  \multicolumn{7}{>{\columncolor[gray]{.95}}c}{\textbf{\textit{Baselines with retrieval}}} \\
 \midrule 
    LLaMA2\textsubscript{{7B}} & 27.8 & 47.8 & 39.8  & 28.5  & 13.6 & 9.59  \\
    LLaMA2\textsubscript{{13B}} & 34.0 & 48.1 & 35.2 & 26.8  & 21.8 & 16.3  \\
    LLaMA2\textsubscript{{7B-chat}}  &27.4  & 52.9 & 43.4 & 25.3  & 34.5 & 33.2  \\
    LLaMA2\textsubscript{{13B-chat}} & 32.7 & 53.5  & 53.4  & 26.4  & 39.4 & 38.4  \\
    Vicuna\textsubscript{{7B}} \text{\cite{vicuna2023}}& 28.0 & 55.2 & 62.4 & 24.3 &  45.7 & 40.8  \\
    Vicuna\textsubscript{{13B}} \text{\cite{vicuna2023}} & 35.4 & 56.1  & 60.6 & 27.3 &  51.3 & 50.2  \\
    LLaMA2-FT\textsubscript{{7B}} & 36.8 & 54.4  & 67.5 & 28.5  & 47.2 & 45.4  \\
    ReAct \cite{yao2023react} & - & - & 64.6 & -  & - & -  \\
    RECOMP \cite{xu2024recomp} & 38.4 & - & - & - & - & -  \\
     Self-RAG\textsubscript{{7B}}  \text{\cite{asai2024selfrag}}  & 37.2 & 54.9 & 70.2 & 30.0 & 66.9 & 67.8  \\
    Self-RAG\textsubscript{{13B}} \text{\cite{asai2024selfrag}} & 38.8 & 55.8 & 72.1 & 31.7 & 70.3 & 71.3  \\
    \midrule 
\textsc{Self-Reasoning}\textsubscript{{7B}} & 38.0 & 54.2 &  78.6& 33.9  & 66.3 & 70.8  \\
\textsc{Self-Reasoning}\textsubscript{{13B}} & \textbf{41.4} & \textbf{57.3}  & \textbf{83.9} & \textbf{35.2}  & \textbf{71.2} & \textbf{72.3}  \\
 \midrule 
 GPT-4 & \textcolor{gray}{\textbf{46.6}} & \textcolor{gray}{\textbf{62.5}}  & \textcolor{gray}{\textbf{87.7}} & \textcolor{gray}{\textbf{41.3}}  & \textcolor{gray}{\textbf{75.6}} & \textcolor{gray}{68.5}  \\
  \bottomrule
\end{tabularx}
\caption{Performance comparisons with different baseline models on two short-form QA datasets, a long-form QA dataset, and a fact verification dataset. The numbers with bold black represent the best results excluding GPT-4. The results are averaged over five runs, and presented with standard variance values omitted (all $\leq 2\% $).}
  \label{tab:model_compare}
\end{table*}

\subsection{Datasets and Settings}{\label{dataset}}
To demonstrate the effectiveness of our proposed \textsc{self-reasoning} framework, we conduct an extensive experimental evaluation on two short-form QA datasets (NaturalQuestion \cite{kwiatkowski-etal-2019-natural} and PopQA \cite{mallen-etal-2023-trust}), one long-form QA dataset (ASQA \cite{stelmakh-etal-2022-asqa}), and one fact verification dataset (FEVER \cite{thorne-etal-2018-fever}). 
Detailed descriptions of the datasets can be found in the Appendix. We explore off-the-shelf retrievers. We use DPR \cite{karpukhin-etal-2020-dense} and Contriever-MS MARCO \cite{izacard2021unsupervised} to retrieve the top five documents from Wikipedia.

By default, we use DPR as a retriever for the NQ, as DPR has been fine-tuned on the high-quality NQ data. On the PopQA, where question and answer pairs are created based on Wikipedia in 2022, therefore, for the PopQA, we use the December 2020 preprocessed Wikipedia corpus provided by \cite{izacard2022few} and use Contriever as a retriever. For the ASQA dataset, we use GTR \cite{ni-etal-2022-large} as a retrieval that corresponds to the experimental settings in \cite{gao-etal-2023-enabling}. More settings can be found in the Appendix.

\subsection{Evaluation Metrics}{\label{ref: shortlong}}
We use different evaluation metrics for short-form QA, long-form QA, and fact verification tasks. 

\paragraph{\textbf{Short-form QA metrics.}}
 We report \textit{accuracy} for short-form QA tasks, which is based on whether ground-truth answers are included in the model predictions instead of strictly requiring exact matching, following \citet{mallen-etal-2023-trust,schick2023toolformer}. 

\paragraph{\textbf{Long-form QA metrics.}}
 For long-form QA tasks, we report the \textit{EM recall} as a correctness metric, and the \textit{citation recall} and the \textit{citation precision} for citation quality, which are the same as the metrics in \cite{gao-etal-2023-enabling}. 

\paragraph{\textbf{Fact verification metrics.}}
For the fact verification task, we report the \textit{accuracy} as a metric, which is a three-class classification accuracy, following \citet{thorne-etal-2018-fever}.

\subsection{Baseline Models}

\paragraph{\textbf{Baseline models without retrieval.}}
 We evaluate strong open-source pre-trained LLMs as baseline models. For basic LLMs, we test LLaMA2-7B, LLaMA2-13B \cite{touvron2023llama} and its instruction-tuned chat version LLaMA2-Chat-7B, LLaMA2-Chat-13B.

\paragraph{\textbf{Baseline models with retrieval.} }
First, we benchmark the models using the LLaMA2 and the Vicuna \cite{vicuna2023} series models for baselines. Additionally, for a fair comparison, we also include LLaMA2-FT, where LLaMA2 is fine-tuned on all the training samples generated by GPT-4 except the self-reasoning trajectories. To establish strong baselines, we compare our method against RECOMP \cite{xu2024recomp}, ReAct \cite{yao2023react}, and Self-RAG \cite{asai2024selfrag}, all of which are trained with extra GPT-4 generated samples or external tools. We also compare our framework with GPT-4 \cite{openai2023gpt}. We include categorical comparisons with the baseline models in the Appendix.

\begin{figure*}[htb]
    \centering
    \includegraphics[width=0.98\textwidth]{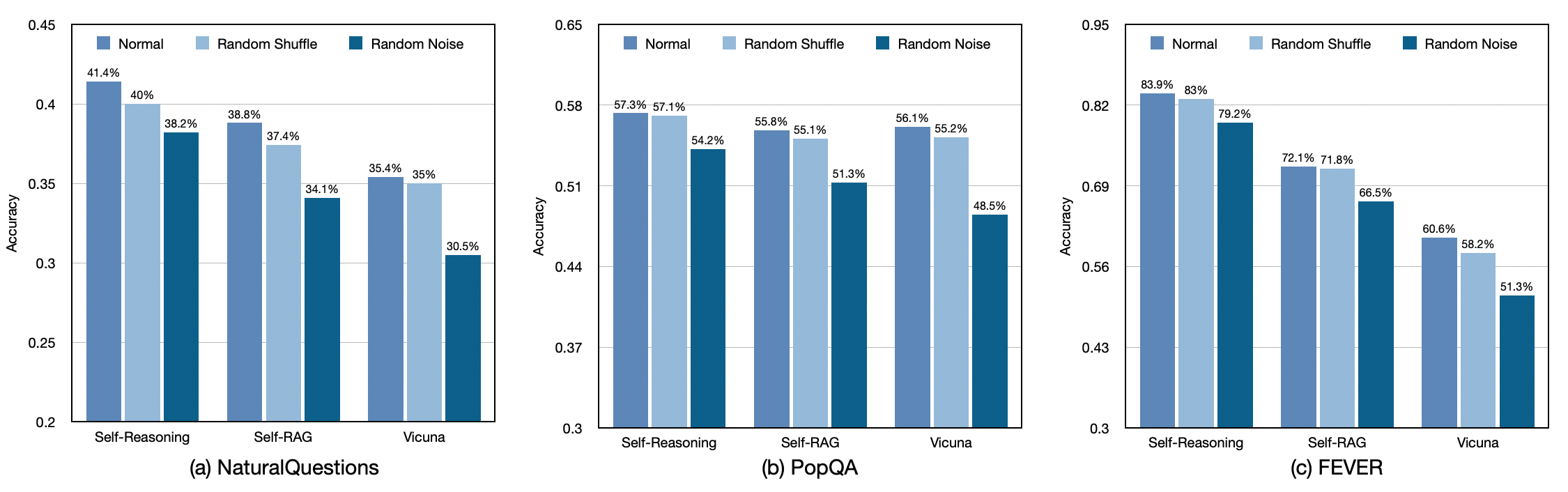}
    \caption{Noise robustness experiment results on three different datasets: (a) On the left is the NQ dataset, (b) in the middle is the PopQA dataset, and (c) on the right is the FEVER dataset. The Self-RAG and Vicuna are 13B parameter size models.}
    \label{fig:robustness}
\end{figure*}

\subsection{Main Results}
Table \ref{tab:model_compare} shows the performance comparisons with different methods on the four public datasets. 
For short-form QA evaluations, the performance of LLMs with augmented retrieval is consistently better than that of basic ones, affirming the effectiveness of the augmented approach.
Notably, under the same order of magnitude parameters, our \textsc{self-reasoning} framework outperforms most of the strong baseline LLMs. Specifically, compared to the Self-RAG, our framework is an end-to-end system trained with only 2,000 self-reasoning trajectory samples. In contrast, the Self-RAG requires training additional critic LMs to predict reflection tokens using an additional 46,000 instances generated by GPT-4. \zhourev{This efficiency not only simplifies the training process but also significantly reduces resource consumption.}

In the context of long-form QA evaluations, for the metrics of \textit{EM recall}, it needs to comprehend multiple documents and merge answers. The EAP and TAP are specifically designed for multi-document reading comprehension, enabling our performance to surpass other baselines. In terms of citation evaluation metrics, the \textsc{self-reasoning} RAG can achieve better results than GPT-4 in ASQA citation recall metrics (72.3 vs. 68.5).
This is largely due to the reasoning trajectories generated in the EAP, which can enhance the recall and precision of citation evaluation, leading to more interpretable and traceable generations.

For fact verification evaluations, we observed that  \textsc{self-seasoning}  is dominantly superior to all baseline models. Our method achieves a much higher accuracy rate than the Self-RAG model (83.9 vs. 72.1). The RAP in our framework is designed to judge the relevance between the retrieved documents and the question, which leads to a notable enhancement in accuracy for this fact verification task.

To clearly demonstrate the practical applications and benefits of our \textsc{self-reasoning} framework, we provide a case study for a more in-depth analysis in Appendix, which illustrates how our framework operates in real-world scenarios.

\begin{table}[t]
\small
\center
  \begin{tabularx}{0.42\textwidth}{p{1.8cm}<{\centering}p{1.2cm}<{\centering}p{1.2cm}<{\centering}p{1.5cm}<{\centering}}
    \toprule
    \multirow{2}{*}{ \textbf{Models} } & \multicolumn{1}{c}{ \textbf{NQ}} & \multicolumn{1}{c}{ \textbf{PopQA} } & \multicolumn{1}{c}{ \textbf{FEVER} } \\
      \cmidrule(r){2-2}  \cmidrule(r){3-3}  \cmidrule(r){4-4} 
& (acc) & (acc) & (acc)  \\
    \midrule
\textsc{Origin}  & 41.4 & 57.3 & 83.9 \\
\hdashline
    
w/o (RAP)  & 39.9 & 54.3 & 72.2 \\
w/o (EAP)  & 37.2 & 53.2 & 78.4  \\
w/o (TAP)  & 38.2 & 53.4 & 81.2  \\
\hdashline
w/o (GL) & 39.5 & 55.3 & 81.2 \\
w/o (QC) & 37.7& 54.2& 80.8 \\

  \bottomrule
\end{tabularx}
\caption{The ablation study on two short-form QA datasets and a fact verification dataset with 13B parameter size models. In the table, the \textsc{origin} represents our self-reasoning model enhanced with self-generated trajectories.}
  \label{tab:ablation}
\end{table}

\begin{table}[t]
\small
\center
  \begin{tabularx}{0.42\textwidth}{p{1.8cm}<{\centering}p{1.2cm}<{\centering}p{1.2cm}<{\centering}p{1.5cm}<{\centering}}
    \toprule

    \multirow{2}{*}{ \textbf{Models} } & \multicolumn{1}{c}{ \textbf{NQ}} & \multicolumn{1}{c}{ \textbf{PopQA} } & \multicolumn{1}{c}{ \textbf{FEVER} } \\
      \cmidrule(r){2-2}  \cmidrule(r){3-3}  \cmidrule(r){4-4} 
& (acc) & (acc) & (acc)  \\
    \midrule
LLaMA2  & 32.7 & 53.5 & 53.4 \\
+ \textit{trajectory}  & 38.3 & 54.2 & 79.2 \\
\hdashline
Vicuna & 35.4 & 56.1 & 60.6  \\
+ \textit{trajectory}  & 38.5 & 56.4 & 79.6  \\
  \bottomrule
\end{tabularx}
\caption{\xia{The analysis on the effectiveness of self-reasoning trajectories with 13B parameter size models. In the table, the \textit{+trajectory} indicates the result of the baseline model is enhanced with self-generated trajectories by our framework.}}
  \label{tab:ablation2}
\end{table}

\section{Analysis}
\subsection{Ablation Study}
We conduct an ablation study on two short-form QA datasets and a fact verification dataset to analyze the individual contributions of each process within our proposed \textsc{self-reasoning} framework. We further explore the effectiveness of the gradual learning (GL) method and the quality control (QC) of data generation (a detailed analysis described in the Appendix). The main ablation study results are shown in Table \ref{tab:ablation} and Table \ref{tab:ablation2}.

\subsubsection*{Effectiveness of RAP.}

First, we evaluate the effect of the RAP. The removal of the RAP causes the overall performance to drop in two short-form QA datasets and a fact verification dataset, suggesting that preliminary consideration of the relevance between questions and retrieved documents can help improve performance. We notice that the performance declines most significantly in the FEVER dataset. Detecting irrelevant documents is critical in the fact-verification task. Our model will immediately output \textit{NotEnoughInfo} if it detects that all documents are irrelevant.

\subsubsection*{Effectiveness of EAP.}
Then we evaluate the effect of the EAP. Removing the EAP causes the overall performance of the average \textit{accuracy} to decline from 60.9 to 56.3 in three short-form QA datasets, which indicates that snippets of key sentences and document citations generated through self-reasoning are instrumental in boosting accuracy.

\subsubsection*{Effectiveness of TAP.}
Finally, we evaluate the effect of the TAP. When excluding the TAP, we can observe a performance decline on all three datasets, demonstrating that self-analysis based on two previous processes generated trajectories can also improve the performance of LLMs. Note that the \textit{analysis} content generated by TAP is indispensable for the long-form QA evaluation.

\subsubsection*{Effectiveness of Self-Reasoning Trajectory.} 
To verify whether the trajectories generated by the self-reasoning framework are truly effective, we put the trajectories generated by our \textsc{self-reasoning} framework into the original baseline models as input prompts, and then use the baseline models to regenerate the answers. We observe that incorporating self-generated trajectories can significantly enhance performance in short QA tasks and fact verification tasks.

\subsection{Retrieval Robustness Analysis}
Retrievers are not perfect and past work has shown that noisy retrieval can have negative effects on the performance of LLMs \cite{petroni2020context, li-etal-2023-large}. In this section, we design two kinds of settings to validate the robustness of RALMs. In the first setting, we test whether the order of the retrieved documents will affect the performance of the RALMs. Specifically, after retrieving the top-$k$ documents using retrievals with a descending relevance score, we randomly shuffle the order of the retrieved documents and then input them to an LLM. In the second setting, we test how noisy documents impact the performance of LLMs. When retrieving the top-$k$ documents from the given question, we randomly replace 50\% of the retrieved documents with other documents sampled from a different question in the dataset.

Figure \ref{fig:robustness} shows the noise robustness experiment results on three datasets. Our \textsc{self-reasoning} framework consistently outperforms the Self-RAG and Vicuna models. We observe that random shuffling of retrieved documents has a minimal impact on the performance of RALMs. If the provided documents are supportive, it is trivial for a RALM to determine the correct answer. However, when presented with noisy documents, all models experience a decline in performance. 
The performance drop in our self-reasoning framework is relatively minimal, demonstrating the robustness of our method even when handling noisy documents.

\begin{figure}[htb]
    \centering
    \includegraphics[width=0.4\textwidth]
    {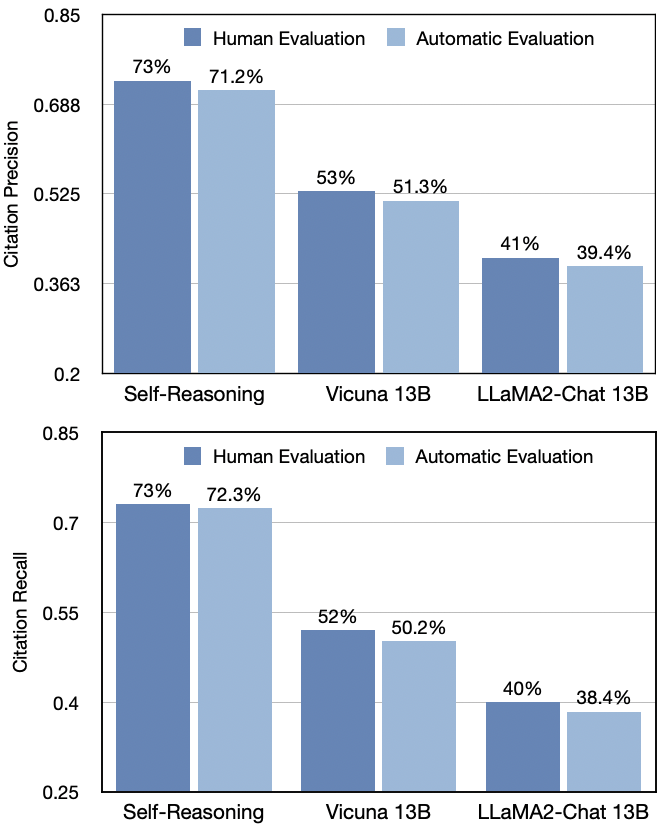}
    \caption{Human citation quality evaluation vs. automatic citation evaluation on the long-form ASQA dataset.}
    \label{fig:ciation}
\end{figure}

\subsection{Citation Analysis}
As the automatic evaluation by the NLI model cannot detect partially supported citations, we discuss the analysis of citations with human evaluation in this section. Similarly to \citet{liu-etal-2023-evaluating}, we conduct a human evaluation on two dimensions: 1) \textit{citation recall}: annotators are given a statement and all documents that the statement refers to and are asked to judge whether the documents fully support the given statement; 2) \textit{citation precision}: given a statement and one of its citations, annotators are asked to validate whether the citation \textit{fully supports}, \textit{partially supports} or \textit{does not support} the statement. 
As shown in Figure \ref{fig:ciation}, the relative rankings by human evaluation align well with those from the automatic evaluation, and the human evaluation often yields a closely higher score when compared with the automatic evaluation. Details of human annotation can be found in the Appendix.

\subsection{Latency Analysis}
We also compared the inference latency of \textsc{self-reasoning} RAG with that of Self-RAG and GPT-4. The results show that our method maintains comparable latency to Self-RAG while delivering substantial performance gains. Detailed results are available in the Appendix.
\section{Conclusion}

\zhou{RALMs can effectively enhance the performance of LLMs in \zhourev{handling} knowledge-intensive tasks.
\zhourev{Despite their effectiveness, notable concerns about their reliability and traceability persist. To address these limitations,} we propose a novel \textsc{self-reasoning} framework to improve the performance of RALMs by using reasoning trajectories generated by the LLM itself. It is comprised of a relevance-aware process, an evidence-aware selective process, and a trajectory analysis process. We conduct extensive experiments on four public datasets to demonstrate the superiority of our framework \zhourev{over} existing state-of-the-art models.}

\bibliography{aaai25,anthology_new}

\begin{thebibliography}{46}
\providecommand{\natexlab}[1]{#1}

\bibitem[{Asai et~al.(2024)Asai, Wu, Wang, Sil, and Hajishirzi}]{asai2024selfrag}
Asai, A.; Wu, Z.; Wang, Y.; Sil, A.; and Hajishirzi, H. 2024.
\newblock Self-{RAG}: Learning to Retrieve, Generate, and Critique through Self-Reflection.
\newblock In \emph{The Twelfth International Conference on Learning Representations}.

\bibitem[{Baek et~al.(2023)Baek, Jeong, Kang, Park, and Hwang}]{baek-etal-2023-knowledge-augmented-language}
Baek, J.; Jeong, S.; Kang, M.; Park, J.; and Hwang, S. 2023.
\newblock Knowledge-Augmented Language Model Verification.
\newblock In Bouamor, H.; Pino, J.; and Bali, K., eds., \emph{Proceedings of the 2023 Conference on Empirical Methods in Natural Language Processing}, 1720--1736. Singapore.

\bibitem[{Borgeaud et~al.(2022)Borgeaud, Mensch, Hoffmann, Cai, Rutherford, Millican, Van Den~Driessche, Lespiau, Damoc, Clark et~al.}]{borgeaud2022improving}
Borgeaud, S.; Mensch, A.; Hoffmann, J.; Cai, T.; Rutherford, E.; Millican, K.; Van Den~Driessche, G.~B.; Lespiau, J.-B.; Damoc, B.; Clark, A.; et~al. 2022.
\newblock Improving language models by retrieving from trillions of tokens.
\newblock In \emph{International conference on machine learning}, 2206--2240. PMLR.

\bibitem[{Brown et~al.(2020)Brown, Mann, Ryder, Subbiah, Kaplan, Dhariwal, Neelakantan, Shyam, Sastry, Askell et~al.}]{brown2020language}
Brown, T.; Mann, B.; Ryder, N.; Subbiah, M.; Kaplan, J.~D.; Dhariwal, P.; Neelakantan, A.; Shyam, P.; Sastry, G.; Askell, A.; et~al. 2020.
\newblock Language models are few-shot learners.
\newblock \emph{Advances in neural information processing systems}, 33: 1877--1901.

\bibitem[{Chiang et~al.(2023)Chiang, Li, Lin, Sheng, Wu, Zhang, Zheng, Zhuang, Zhuang, Gonzalez, Stoica, and Xing}]{vicuna2023}
Chiang, W.-L.; Li, Z.; Lin, Z.; Sheng, Y.; Wu, Z.; Zhang, H.; Zheng, L.; Zhuang, S.; Zhuang, Y.; Gonzalez, J.~E.; Stoica, I.; and Xing, E.~P. 2023.
\newblock Vicuna: An Open-Source Chatbot Impressing GPT-4 with 90\%* ChatGPT Quality.

\bibitem[{Gao et~al.(2023{\natexlab{a}})Gao, Dai, Pasupat, Chen, Chaganty, Fan, Zhao, Lao, Lee, Juan, and Guu}]{gao-etal-2023-rarr}
Gao, L.; Dai, Z.; Pasupat, P.; Chen, A.; Chaganty, A.~T.; Fan, Y.; Zhao, V.; Lao, N.; Lee, H.; Juan, D.-C.; and Guu, K. 2023{\natexlab{a}}.
\newblock {RARR}: Researching and Revising What Language Models Say, Using Language Models.
\newblock In Rogers, A.; Boyd-Graber, J.; and Okazaki, N., eds., \emph{Proceedings of the 61st Annual Meeting of the Association for Computational Linguistics (Volume 1: Long Papers)}, 16477--16508. Toronto, Canada.

\bibitem[{Gao et~al.(2023{\natexlab{b}})Gao, Yen, Yu, and Chen}]{gao-etal-2023-enabling}
Gao, T.; Yen, H.; Yu, J.; and Chen, D. 2023{\natexlab{b}}.
\newblock Enabling Large Language Models to Generate Text with Citations.
\newblock In Bouamor, H.; Pino, J.; and Bali, K., eds., \emph{Proceedings of the 2023 Conference on Empirical Methods in Natural Language Processing}, 6465--6488. Singapore.

\bibitem[{Guu et~al.(2020)Guu, Lee, Tung, Pasupat, and Chang}]{10.5555/3524938.3525306}
Guu, K.; Lee, K.; Tung, Z.; Pasupat, P.; and Chang, M.-W. 2020.
\newblock REALM: retrieval-augmented language model pre-training.
\newblock In \emph{Proceedings of the 37th International Conference on Machine Learning}, ICML'20. JMLR.org.

\bibitem[{Honovich et~al.(2022)Honovich, Aharoni, Herzig, Taitelbaum, Kukliansy, Cohen, Scialom, Szpektor, Hassidim, and Matias}]{honovich2022true}
Honovich, O.; Aharoni, R.; Herzig, J.; Taitelbaum, H.; Kukliansy, D.; Cohen, V.; Scialom, T.; Szpektor, I.; Hassidim, A.; and Matias, Y. 2022.
\newblock TRUE: Re-evaluating factual consistency evaluation.
\newblock \emph{arXiv preprint arXiv:2204.04991}.

\bibitem[{Izacard et~al.(2021)Izacard, Caron, Hosseini, Riedel, Bojanowski, Joulin, and Grave}]{izacard2021unsupervised}
Izacard, G.; Caron, M.; Hosseini, L.; Riedel, S.; Bojanowski, P.; Joulin, A.; and Grave, E. 2021.
\newblock Unsupervised dense information retrieval with contrastive learning.
\newblock \emph{arXiv preprint arXiv:2112.09118}.

\bibitem[{Izacard et~al.(2022)Izacard, Lewis, Lomeli, Hosseini, Petroni, Schick, Dwivedi-Yu, Joulin, Riedel, and Grave}]{izacard2022few}
Izacard, G.; Lewis, P.; Lomeli, M.; Hosseini, L.; Petroni, F.; Schick, T.; Dwivedi-Yu, J.; Joulin, A.; Riedel, S.; and Grave, E. 2022.
\newblock Few-shot learning with retrieval augmented language models.
\newblock \emph{arXiv preprint arXiv:2208.03299}.

\bibitem[{Ji et~al.(2023)Ji, Lee, Frieske, Yu, Su, Xu, Ishii, Bang, Madotto, and Fung}]{ji2023survey}
Ji, Z.; Lee, N.; Frieske, R.; Yu, T.; Su, D.; Xu, Y.; Ishii, E.; Bang, Y.~J.; Madotto, A.; and Fung, P. 2023.
\newblock Survey of hallucination in natural language generation.
\newblock \emph{ACM Computing Surveys}, 55(12): 1--38.

\bibitem[{Jiang et~al.(2023)Jiang, Xu, Gao, Sun, Liu, Dwivedi-Yu, Yang, Callan, and Neubig}]{jiang-etal-2023-active}
Jiang, Z.; Xu, F.; Gao, L.; Sun, Z.; Liu, Q.; Dwivedi-Yu, J.; Yang, Y.; Callan, J.; and Neubig, G. 2023.
\newblock Active Retrieval Augmented Generation.
\newblock In Bouamor, H.; Pino, J.; and Bali, K., eds., \emph{Proceedings of the 2023 Conference on Empirical Methods in Natural Language Processing}, 7969--7992. Singapore.

\bibitem[{Karpukhin et~al.(2020)Karpukhin, Oguz, Min, Lewis, Wu, Edunov, Chen, and Yih}]{karpukhin-etal-2020-dense}
Karpukhin, V.; Oguz, B.; Min, S.; Lewis, P.; Wu, L.; Edunov, S.; Chen, D.; and Yih, W.-t. 2020.
\newblock Dense Passage Retrieval for Open-Domain Question Answering.
\newblock In Webber, B.; Cohn, T.; He, Y.; and Liu, Y., eds., \emph{Proceedings of the 2020 Conference on Empirical Methods in Natural Language Processing (EMNLP)}, 6769--6781. Online.

\bibitem[{Kwiatkowski et~al.(2019)Kwiatkowski, Palomaki, Redfield, Collins, Parikh, Alberti, Epstein, Polosukhin, Devlin, Lee, Toutanova, Jones, Kelcey, Chang, Dai, Uszkoreit, Le, and Petrov}]{kwiatkowski-etal-2019-natural}
Kwiatkowski, T.; Palomaki, J.; Redfield, O.; Collins, M.; Parikh, A.; Alberti, C.; Epstein, D.; Polosukhin, I.; Devlin, J.; Lee, K.; Toutanova, K.; Jones, L.; Kelcey, M.; Chang, M.-W.; Dai, A.~M.; Uszkoreit, J.; Le, Q.; and Petrov, S. 2019.
\newblock Natural Questions: A Benchmark for Question Answering Research.
\newblock \emph{Transactions of the Association for Computational Linguistics}, 7: 452--466.

\bibitem[{Kwon et~al.(2023)Kwon, Li, Zhuang, Sheng, Zheng, Yu, Gonzalez, Zhang, and Stoica}]{kwon2023efficient}
Kwon, W.; Li, Z.; Zhuang, S.; Sheng, Y.; Zheng, L.; Yu, C.~H.; Gonzalez, J.~E.; Zhang, H.; and Stoica, I. 2023.
\newblock Efficient Memory Management for Large Language Model Serving with PagedAttention.
\newblock In \emph{Proceedings of the ACM SIGOPS 29th Symposium on Operating Systems Principles}.

\bibitem[{Lewis et~al.(2020)Lewis, Perez, Piktus, Petroni, Karpukhin, Goyal, K{\"u}ttler, Lewis, Yih, Rockt{\"a}schel et~al.}]{lewis2020retrieval}
Lewis, P.; Perez, E.; Piktus, A.; Petroni, F.; Karpukhin, V.; Goyal, N.; K{\"u}ttler, H.; Lewis, M.; Yih, W.-t.; Rockt{\"a}schel, T.; et~al. 2020.
\newblock Retrieval-augmented generation for knowledge-intensive nlp tasks.
\newblock \emph{Advances in Neural Information Processing Systems}, 33: 9459--9474.

\bibitem[{Li et~al.(2023)Li, Rawat, Zaheer, Wang, Lukasik, Veit, Yu, and Kumar}]{li-etal-2023-large}
Li, D.; Rawat, A.~S.; Zaheer, M.; Wang, X.; Lukasik, M.; Veit, A.; Yu, F.; and Kumar, S. 2023.
\newblock Large Language Models with Controllable Working Memory.
\newblock In Rogers, A.; Boyd-Graber, J.; and Okazaki, N., eds., \emph{Findings of the Association for Computational Linguistics: ACL 2023}, 1774--1793. Toronto, Canada.

\bibitem[{Liu, Zhang, and Liang(2023)}]{liu-etal-2023-evaluating}
Liu, N.; Zhang, T.; and Liang, P. 2023.
\newblock Evaluating Verifiability in Generative Search Engines.
\newblock In Bouamor, H.; Pino, J.; and Bali, K., eds., \emph{Findings of the Association for Computational Linguistics: EMNLP 2023}, 7001--7025. Singapore.

\bibitem[{Mallen et~al.(2023)Mallen, Asai, Zhong, Das, Khashabi, and Hajishirzi}]{mallen-etal-2023-trust}
Mallen, A.; Asai, A.; Zhong, V.; Das, R.; Khashabi, D.; and Hajishirzi, H. 2023.
\newblock When Not to Trust Language Models: Investigating Effectiveness of Parametric and Non-Parametric Memories.
\newblock In Rogers, A.; Boyd-Graber, J.; and Okazaki, N., eds., \emph{Proceedings of the 61st Annual Meeting of the Association for Computational Linguistics (Volume 1: Long Papers)}, 9802--9822. Toronto, Canada.

\bibitem[{Menick et~al.(2022)Menick, Trebacz, Mikulik, Aslanides, Song, Chadwick, Glaese, Young, Campbell-Gillingham, Irving et~al.}]{menick2022teaching}
Menick, J.; Trebacz, M.; Mikulik, V.; Aslanides, J.; Song, F.; Chadwick, M.; Glaese, M.; Young, S.; Campbell-Gillingham, L.; Irving, G.; et~al. 2022.
\newblock Teaching language models to support answers with verified quotes.
\newblock \emph{arXiv preprint arXiv:2203.11147}.

\bibitem[{Min et~al.(2020)Min, Michael, Hajishirzi, and Zettlemoyer}]{min-etal-2020-ambigqa}
Min, S.; Michael, J.; Hajishirzi, H.; and Zettlemoyer, L. 2020.
\newblock {A}mbig{QA}: Answering Ambiguous Open-domain Questions.
\newblock In Webber, B.; Cohn, T.; He, Y.; and Liu, Y., eds., \emph{Proceedings of the 2020 Conference on Empirical Methods in Natural Language Processing (EMNLP)}, 5783--5797. Online.

\bibitem[{Nakano et~al.(2021)Nakano, Hilton, Balaji, Wu, Ouyang, Kim, Hesse, Jain, Kosaraju, Saunders et~al.}]{nakano2021webgpt}
Nakano, R.; Hilton, J.; Balaji, S.; Wu, J.; Ouyang, L.; Kim, C.; Hesse, C.; Jain, S.; Kosaraju, V.; Saunders, W.; et~al. 2021.
\newblock Webgpt: Browser-assisted question-answering with human feedback.
\newblock \emph{arXiv preprint arXiv:2112.09332}.

\bibitem[{Ni et~al.(2022)Ni, Qu, Lu, Dai, Hernandez~Abrego, Ma, Zhao, Luan, Hall, Chang, and Yang}]{ni-etal-2022-large}
Ni, J.; Qu, C.; Lu, J.; Dai, Z.; Hernandez~Abrego, G.; Ma, J.; Zhao, V.; Luan, Y.; Hall, K.; Chang, M.-W.; and Yang, Y. 2022.
\newblock Large Dual Encoders Are Generalizable Retrievers.
\newblock In Goldberg, Y.; Kozareva, Z.; and Zhang, Y., eds., \emph{Proceedings of the 2022 Conference on Empirical Methods in Natural Language Processing}, 9844--9855. Abu Dhabi, United Arab Emirates.

\bibitem[{OpenAI(2023)}]{openai2023gpt}
OpenAI, R. 2023.
\newblock Gpt-4 technical report. arxiv 2303.08774.
\newblock \emph{View in Article}, 2: 13.

\bibitem[{Pan et~al.(2024)Pan, Luo, Li, and Liu}]{pan2024chain}
Pan, Z.; Luo, H.; Li, M.; and Liu, H. 2024.
\newblock Chain-of-action: Faithful and multimodal question answering through large language models.
\newblock \emph{arXiv preprint arXiv:2403.17359}.

\bibitem[{Petroni et~al.(2020)Petroni, Lewis, Piktus, Rockt{\"a}schel, Wu, Miller, and Riedel}]{petroni2020context}
Petroni, F.; Lewis, P.; Piktus, A.; Rockt{\"a}schel, T.; Wu, Y.; Miller, A.~H.; and Riedel, S. 2020.
\newblock How context affects language models' factual predictions.
\newblock \emph{arXiv preprint arXiv:2005.04611}.

\bibitem[{Petroni et~al.(2021)Petroni, Piktus, Fan, Lewis, Yazdani, De~Cao, Thorne, Jernite, Karpukhin, Maillard, Plachouras, Rockt{\"a}schel, and Riedel}]{petroni-etal-2021-kilt}
Petroni, F.; Piktus, A.; Fan, A.; Lewis, P.; Yazdani, M.; De~Cao, N.; Thorne, J.; Jernite, Y.; Karpukhin, V.; Maillard, J.; Plachouras, V.; Rockt{\"a}schel, T.; and Riedel, S. 2021.
\newblock {KILT}: a Benchmark for Knowledge Intensive Language Tasks.
\newblock In Toutanova, K.; Rumshisky, A.; Zettlemoyer, L.; Hakkani-Tur, D.; Beltagy, I.; Bethard, S.; Cotterell, R.; Chakraborty, T.; and Zhou, Y., eds., \emph{Proceedings of the 2021 Conference of the North American Chapter of the Association for Computational Linguistics: Human Language Technologies}, 2523--2544. Online.

\bibitem[{Press et~al.(2023)Press, Zhang, Min, Schmidt, Smith, and Lewis}]{press-etal-2023-measuring}
Press, O.; Zhang, M.; Min, S.; Schmidt, L.; Smith, N.; and Lewis, M. 2023.
\newblock Measuring and Narrowing the Compositionality Gap in Language Models.
\newblock In Bouamor, H.; Pino, J.; and Bali, K., eds., \emph{Findings of the Association for Computational Linguistics: EMNLP 2023}, 5687--5711. Singapore.

\bibitem[{Ram et~al.(2023)Ram, Levine, Dalmedigos, Muhlgay, Shashua, Leyton-Brown, and Shoham}]{ram-etal-2023-context}
Ram, O.; Levine, Y.; Dalmedigos, I.; Muhlgay, D.; Shashua, A.; Leyton-Brown, K.; and Shoham, Y. 2023.
\newblock In-Context Retrieval-Augmented Language Models.
\newblock \emph{Transactions of the Association for Computational Linguistics}, 11: 1316--1331.

\bibitem[{Rasley et~al.(2020)Rasley, Rajbhandari, Ruwase, and He}]{rasley2020deepspeed}
Rasley, J.; Rajbhandari, S.; Ruwase, O.; and He, Y. 2020.
\newblock Deepspeed: System optimizations enable training deep learning models with over 100 billion parameters.
\newblock In \emph{Proceedings of the 26th ACM SIGKDD International Conference on Knowledge Discovery \& Data Mining}, 3505--3506.

\bibitem[{Schick et~al.(2023)Schick, Dwivedi-Yu, Dess{\`\i}, Raileanu, Lomeli, Zettlemoyer, Cancedda, and Scialom}]{schick2023toolformer}
Schick, T.; Dwivedi-Yu, J.; Dess{\`\i}, R.; Raileanu, R.; Lomeli, M.; Zettlemoyer, L.; Cancedda, N.; and Scialom, T. 2023.
\newblock Toolformer: Language models can teach themselves to use tools.
\newblock \emph{arXiv preprint arXiv:2302.04761}.

\bibitem[{Stelmakh et~al.(2022)Stelmakh, Luan, Dhingra, and Chang}]{stelmakh-etal-2022-asqa}
Stelmakh, I.; Luan, Y.; Dhingra, B.; and Chang, M.-W. 2022.
\newblock {ASQA}: Factoid Questions Meet Long-Form Answers.
\newblock In Goldberg, Y.; Kozareva, Z.; and Zhang, Y., eds., \emph{Proceedings of the 2022 Conference on Empirical Methods in Natural Language Processing}, 8273--8288. Abu Dhabi, United Arab Emirates.

\bibitem[{Thorne et~al.(2018)Thorne, Vlachos, Christodoulopoulos, and Mittal}]{thorne-etal-2018-fever}
Thorne, J.; Vlachos, A.; Christodoulopoulos, C.; and Mittal, A. 2018.
\newblock {FEVER}: a Large-scale Dataset for Fact Extraction and {VER}ification.
\newblock In Walker, M.; Ji, H.; and Stent, A., eds., \emph{Proceedings of the 2018 Conference of the North {A}merican Chapter of the Association for Computational Linguistics: Human Language Technologies, Volume 1 (Long Papers)}, 809--819. New Orleans, Louisiana.

\bibitem[{Touvron et~al.(2023)Touvron, Martin, Stone, Albert, Almahairi, Babaei, Bashlykov, Batra, Bhargava, Bhosale et~al.}]{touvron2023llama}
Touvron, H.; Martin, L.; Stone, K.; Albert, P.; Almahairi, A.; Babaei, Y.; Bashlykov, N.; Batra, S.; Bhargava, P.; Bhosale, S.; et~al. 2023.
\newblock Llama 2: Open foundation and fine-tuned chat models.
\newblock \emph{arXiv preprint arXiv:2307.09288}.

\bibitem[{Trivedi et~al.(2023)Trivedi, Balasubramanian, Khot, and Sabharwal}]{trivedi2023interleaving}
Trivedi, H.; Balasubramanian, N.; Khot, T.; and Sabharwal, A. 2023.
\newblock Interleaving Retrieval with Chain-of-Thought Reasoning for Knowledge-Intensive Multi-Step Questions.
\newblock In \emph{Proceedings of the 61st Annual Meeting of the Association for Computational Linguistics (Volume 1: Long Papers)}, 10014--10037.

\bibitem[{Wang et~al.(2022)Wang, Wei, Schuurmans, Le, hsin Chi, and Zhou}]{Wang2022SelfConsistencyIC}
Wang, X.; Wei, J.; Schuurmans, D.; Le, Q.; hsin Chi, E.~H.; and Zhou, D. 2022.
\newblock Self-Consistency Improves Chain of Thought Reasoning in Language Models.
\newblock \emph{ArXiv}, abs/2203.11171.

\bibitem[{Wei et~al.(2022)Wei, Wang, Schuurmans, Bosma, Xia, Chi, Le, Zhou et~al.}]{wei2022chain}
Wei, J.; Wang, X.; Schuurmans, D.; Bosma, M.; Xia, F.; Chi, E.; Le, Q.~V.; Zhou, D.; et~al. 2022.
\newblock Chain-of-thought prompting elicits reasoning in large language models.
\newblock \emph{Advances in neural information processing systems}, 35: 24824--24837.

\bibitem[{Xu et~al.(2024)}]{xu2024recomp}
Xu, F.; et~al. 2024.
\newblock {RECOMP}: Improving Retrieval-Augmented {LM}s with Context Compression and Selective Augmentation.
\newblock In \emph{The Twelfth International Conference on Learning Representations}.

\bibitem[{Xu et~al.(2023)Xu, Pang, Shen, Cheng, and Chua}]{xu2023search}
Xu, S.; Pang, L.; Shen, H.; Cheng, X.; and Chua, T.-s. 2023.
\newblock Search-in-the-chain: Towards the accurate, credible and traceable content generation for complex knowledge-intensive tasks.
\newblock \emph{arXiv preprint arXiv:2304.14732}.

\bibitem[{Yan et~al.(2024)Yan, Gu, Zhu, and Ling}]{yan2024corrective}
Yan, S.-Q.; Gu, J.-C.; Zhu, Y.; and Ling, Z.-H. 2024.
\newblock Corrective retrieval augmented generation.
\newblock \emph{arXiv preprint arXiv:2401.15884}.

\bibitem[{Yao et~al.(2023)Yao, Zhao, Yu, Du, Shafran, Narasimhan, and Cao}]{yao2023react}
Yao, S.; Zhao, J.; Yu, D.; Du, N.; Shafran, I.; Narasimhan, K.; and Cao, Y. 2023.
\newblock {ReAct}: Synergizing Reasoning and Acting in Language Models.
\newblock In \emph{International Conference on Learning Representations (ICLR)}.

\bibitem[{Yoran et~al.(2023)Yoran, Wolfson, Ram, and Berant}]{yoran2023making}
Yoran, O.; Wolfson, T.; Ram, O.; and Berant, J. 2023.
\newblock Making retrieval-augmented language models robust to irrelevant context.
\newblock \emph{arXiv preprint arXiv:2310.01558}.

\bibitem[{Yu et~al.(2023)Yu, Zhang, Pan, Ma, Wang, and Yu}]{yu2023chain}
Yu, W.; Zhang, H.; Pan, X.; Ma, K.; Wang, H.; and Yu, D. 2023.
\newblock Chain-of-note: Enhancing robustness in retrieval-augmented language models.
\newblock \emph{arXiv preprint arXiv:2311.09210}.

\bibitem[{Zhou et~al.(2023)Zhou, Liu, Xu, Iyer, Sun, Mao, Ma, Efrat, Yu, Yu et~al.}]{zhou2023lima}
Zhou, C.; Liu, P.; Xu, P.; Iyer, S.; Sun, J.; Mao, Y.; Ma, X.; Efrat, A.; Yu, P.; Yu, L.; et~al. 2023.
\newblock Lima: Less is more for alignment.
\newblock \emph{arXiv preprint arXiv:2305.11206}.

\bibitem[{Zhou et~al.(2022)Zhou, Scharli, Hou, Wei, Scales, Wang, Schuurmans, Bousquet, Le, and hsin Chi}]{Zhou2022LeasttoMostPE}
Zhou, D.; Scharli, N.; Hou, L.; Wei, J.; Scales, N.; Wang, X.; Schuurmans, D.; Bousquet, O.; Le, Q.; and hsin Chi, E.~H. 2022.
\newblock Least-to-Most Prompting Enables Complex Reasoning in Large Language Models.
\newblock \emph{ArXiv}, abs/2205.10625.

\end{thebibliography}

\newpage

\appendix

\clearpage
\section{{Appendix}}

\begin{table*}[htb]
\small
\center
\begin{tabularx}{0.95\textwidth}{lccccc}

 \toprule 
 
 \multirow{2}{*}{ \textbf{Models} } & \multirow{2}{*}{ \textbf{End-to-End} } &\multicolumn{3}{c}{\textbf{External module}} & \multirow{2}{*}{\textbf{Train data for LLM}}  \\
  \cmidrule(l){3-5} 
&  & (train) & (inference) & (data) &    \\

 \midrule 
Self-Reasoning (Ours) & Y & N & N &  No Need & 2K \\
Self-RAG \cite{asai2024selfrag} & N & N & N & No Need & 145K (Generator)/ 46K (Critic)  \\
ReAct \cite{yao2023react} & N & N & Y & No Need & No Need   \\
RECOMP \cite{xu2024recomp} & N & Y & Y & 152K & No Need   \\
  \bottomrule
\end{tabularx}
\caption{Categorical comparisons with strong baseline models. \textit{External module (train)} and \textit{External module (data)} refer to whether the external module needs to be trained and the number of samples required, respectively. \textit{External module (inference)} indicates whether the external module is needed during the inference stage. \textit{Train data for LLM} indicates the number of training samples needed to train with LLMs.}
  \label{tab:app}
\end{table*}

\subsection{More Related Work of LMs for Reasoning}{\label{app relate work}}

One of the most well-known methods of using LLMs for reasoning is the Chain-of-Thought (CoT) \cite{wei2022chain}, which demonstrates the capability of LLMs to create their thinking process for problem-solving. \citet{Zhou2022LeasttoMostPE} proposes a least-to-most prompting for solving complex tasks. \citet{Wang2022SelfConsistencyIC} introduces a method to reason with self-consistency. \citet{press-etal-2023-measuring} proposes a method to further improve the chain of thought by reasoning explicitly instead of implicitly. 

Recent works have extended beyond the internal reasoning ability of LLMs to include interactions with external tools (e.g., search engines or retrievers) for solving complex tasks. The ReAct \cite{yao2023react} presents an iterative paradigm to combine reasoning and acting with LLMs for tackling language reasoning and decision-making tasks. \citet{xu2023search} introduces a framework to enable information retrieval and LLMs to interact with each other effectively with chain-of-query decomposition. \citet{pan2024chain} proposes a novel framework named Chain-of-Action (CoA), which integrates a reasoning retrieval method to decompose complex questions into chains of configurable actions. 

Different from the above works, which are mostly based on relatively large LLMs (e.g., ChatGPT), our proposed method focuses on enhancing smaller LLMs (e.g., LLaMA2) using only a limited number of samples to achieve high robustness and interpretability through single-step interaction.

\subsection{Instructions}
\begin{figure*}
    \center
    \includegraphics[width=0.95\linewidth]{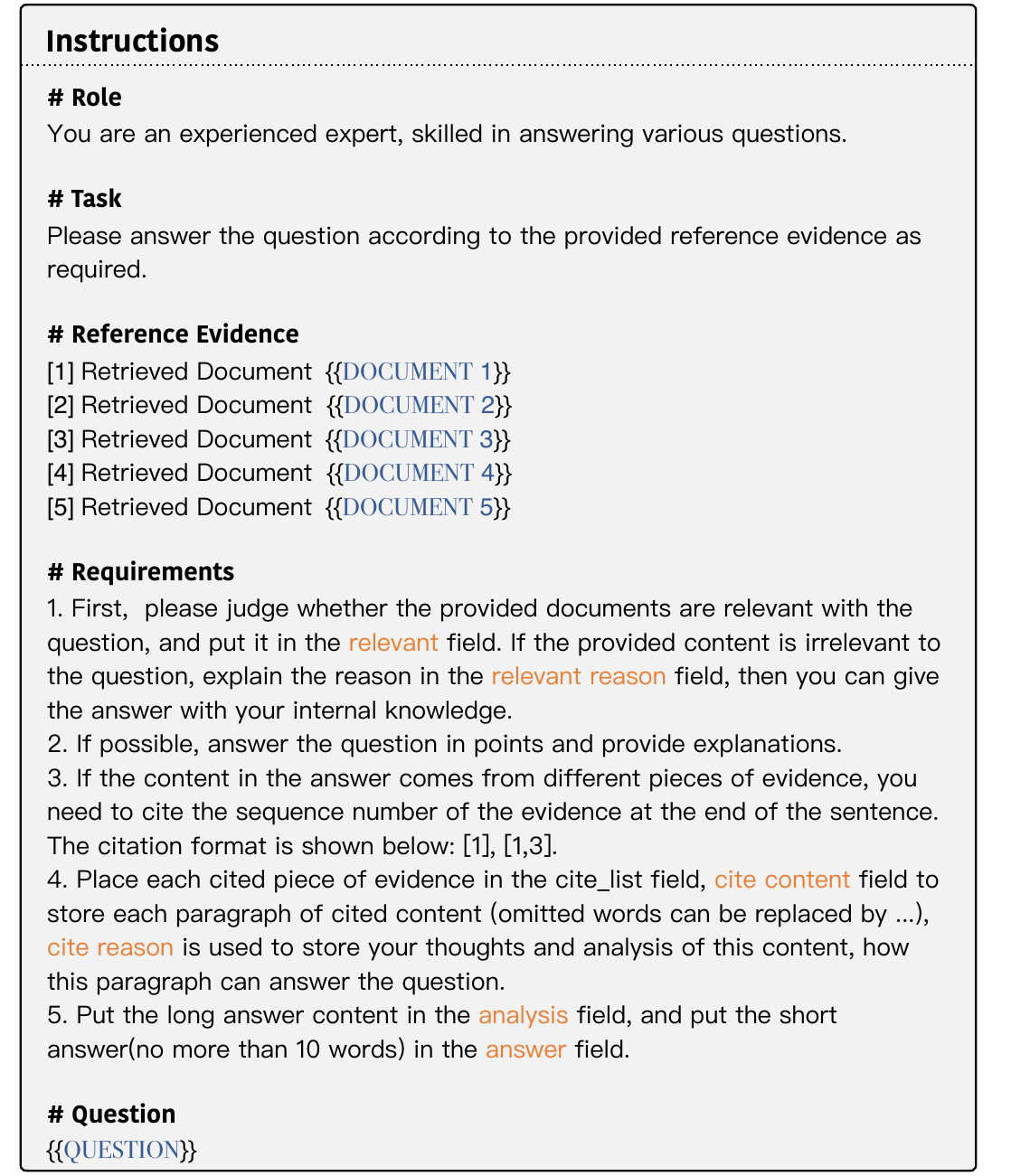}
    \caption{The instructions for the GPT-4 to generate the self-reasoning trajectories for short-form and long-form QA tasks.}
    \label{fig:instruct gpt4}
\end{figure*}

\begin{figure*}
    \center
    \includegraphics[width=0.95\linewidth]{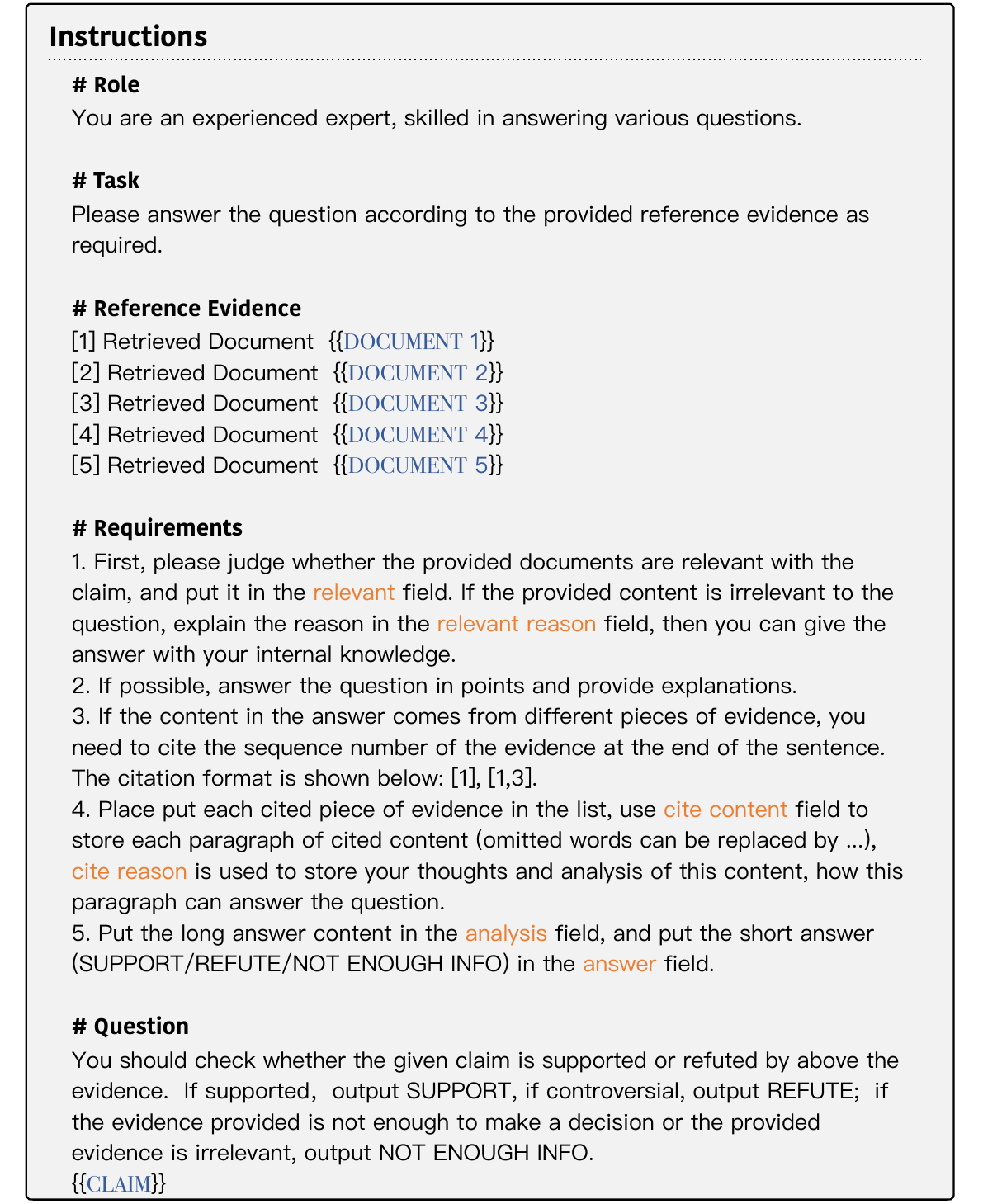}
    \caption{The instructions for the GPT-4 to generate the self-reasoning trajectories for the fact verification task.}
    \label{fig:instruct gpt4 fever}
\end{figure*}

{\label{instruction}}
The instructions for GPT-4 to generate self-reasoning trajectories are shown in Figure \ref{fig:instruct gpt4} (the short-form and long-form QA tasks) and Figure \ref{fig:instruct gpt4 fever} (the fact verification task). The words in the orange font are key fields that need to be generated.

\subsection{Datasets Description}{\label{app:dataset}}
We conducted an extensive experimental evaluation of two short-form QA datasets, one long-form QA dataset, and a fact verification dataset.

\textbf{NaturalQuestion (NQ)} \cite{kwiatkowski-etal-2019-natural} contains real user questions issued to the Google search and answers found from Wikipedia by the annotators. NQ is created to train and evaluate automated question answering systems.

\textbf{PopQA} \cite{mallen-etal-2023-trust} is a large-scale open-domain question answering dataset, consisting of entity-centric QA pairs. Each question is made by converting a knowledge triplet retrieved from Wikidata using a template. In this work, we use PopQA to evaluate performance in long-tail settings.

\textbf{ASQA} \cite{stelmakh-etal-2022-asqa} is a long-form factoid dataset, and most questions can be answered by Wikipedia. 
 Each question originates from AmbigQA \cite{min-etal-2020-ambigqa} and represents an ambiguous query that requires multiple short answers to cover various aspects. The dataset provides a long-form answer that contains all short answers.
 
\textbf{FEVER} \cite{thorne-etal-2018-fever} is a fact verification dataset that contains claims generated by rewriting sentences extracted from Wikipedia and subsequently verified without knowledge of the sentence from which they were derived. The claims are classified as \textit{Supported}, \textit{Refuted}, or \textit{NotEnoughInfo}.

\subsection{Experiment settings}{\label{hyper}}
\label{sec:appendix}

\paragraph{Training settings. }
During gradual learning, we fine-tune the LLaMA-2 \cite{touvron2023llama} model with our self-reasoning framework for 3 epochs with a batch size set to 32, leveraging the DeepSpeed library \cite{rasley2020deepspeed} and the ZeRO optimizer, and we use parameter partitioning ZeRO stage 3 with float16 precision. The learning rate $r_{a}$ for the first stage is set to 5e-5, the learning rate $r_{b}$ for the second stage is set to 3e-5, and the learning rate $r_{c}$ for the final stage is set to 1e-5. Our \textsc{self-reasoning} 13B model is trained on the NVIDIA Tesla 8 $\times$ V100 32GB GPU for 4 hours, while the 7B model is trained for 2 hours. 

\paragraph{Inference settings. }
We use the vLLM $\footnote{Codes are available at
\url{https://github.com/vllm-project/vllm}}$ framework \cite{kwon2023efficient} to accelerate the inference speed during inference. The codes follow the Apache-2.0 license agreement. We use greedy decoding in all experiments to ensure deterministic generations. We test the temperature within a range of $\{0.2, 0.4, 0.6, 0.8, 1.0\}$, finally we set the temperature to 0.2, as we observed lower temperature results in better performance in the open-domain question answering task. The maximum generation length is set to 2048 for our model. All baseline models are tested with zero-shot settings for short-form QA datasets, and with one-shot settings for the long-form QA and fact verification datasets.

\paragraph{Other settings. }
For the document retrieval, we retrieve the top-$k$ relevant documents, and the $k$ is set to 5. We use the DPR and the Contriever in short-form QA settings. For long-form QA, we use GTR as a retrieval and evaluate it using one-shot to instruct the model to generate citations. For the data generation quality control setting, the threshold for citation recall is set to 0.8, and the threshold of citation precision is set to 0.8.

\subsection{Categorical Comparisons}{\label{app:comp}}
We differentiate our method from existing strong baseline models by categorizing and comparing it across five dimensions, as presented in Table \ref{tab:app}. As illustrated in the table, our method can stand out in several key aspects.

First, our \textsc{self-reasoning} method is the only end-to-end framework among existing methods that can improve performance without \zhourev{relying on} external models or tools. Second, our method eliminates the need for external modules during both the training and inference phases. In practical applications, our framework does not need to call multiple tools or modules. Third, our framework requires a significantly smaller dataset for training the LLM compared to other methods, needing only 2,000 samples with self-reasoning trajectories. \zhourev{This efficiency in training drastically lowers the resources and time needed, making our method both cost-effective and scalable for practical applications.}

\begin{algorithm}[t]  
  \caption{Data Quality Control}  
  \label{data:qc}  
  \begin{small}
  \begin{algorithmic}[1]  
    \Require  
       the origin self-reasoning dataset $\mathcal{D}_0$ generated by GPT-4 
    \Ensure  
      filtered high quality self-reasoning dataset $\mathcal{D}_{sr}$
      
    \State Initialize the evaluation metrics program $P$
    \State Initialize the citation score tools $T$ \cite{gao-etal-2023-enabling}
        \For {$i=1$ to $N$} 
            \State Evaluate whether the answer of sample $d_i$ is correct using program $P$
            \If {True and $d_i \in $ Long-form QA dataset }
                \State Compute citation recall score $s_r$ for sample $d_i$ 
                \State Compute citation precision score $s_p$ for sample $d_i$ 
                \If { $s_r \geq \delta_{r}$ and $s_p \geq \delta_{p}$ } 
                    \State Add the training sample to $\mathcal{D}_{l}$
                \EndIf
            \EndIf
            \If {True and $d_i \notin $ Long-form QA dataset }
                 \State Add the training sample to $\mathcal{D}_{s}$
            \EndIf
        \EndFor
    \State $\mathcal{D}_{sr} =  \mathcal{D}_{l} \cup \mathcal{D}_{s}$

    \State \Return  $\mathcal{D}_{sr}$
  \end{algorithmic}
  \end{small}
\end{algorithm}

\subsection{Case Study}
\begin{figure*}
    \center
    \includegraphics[width=0.95\linewidth]{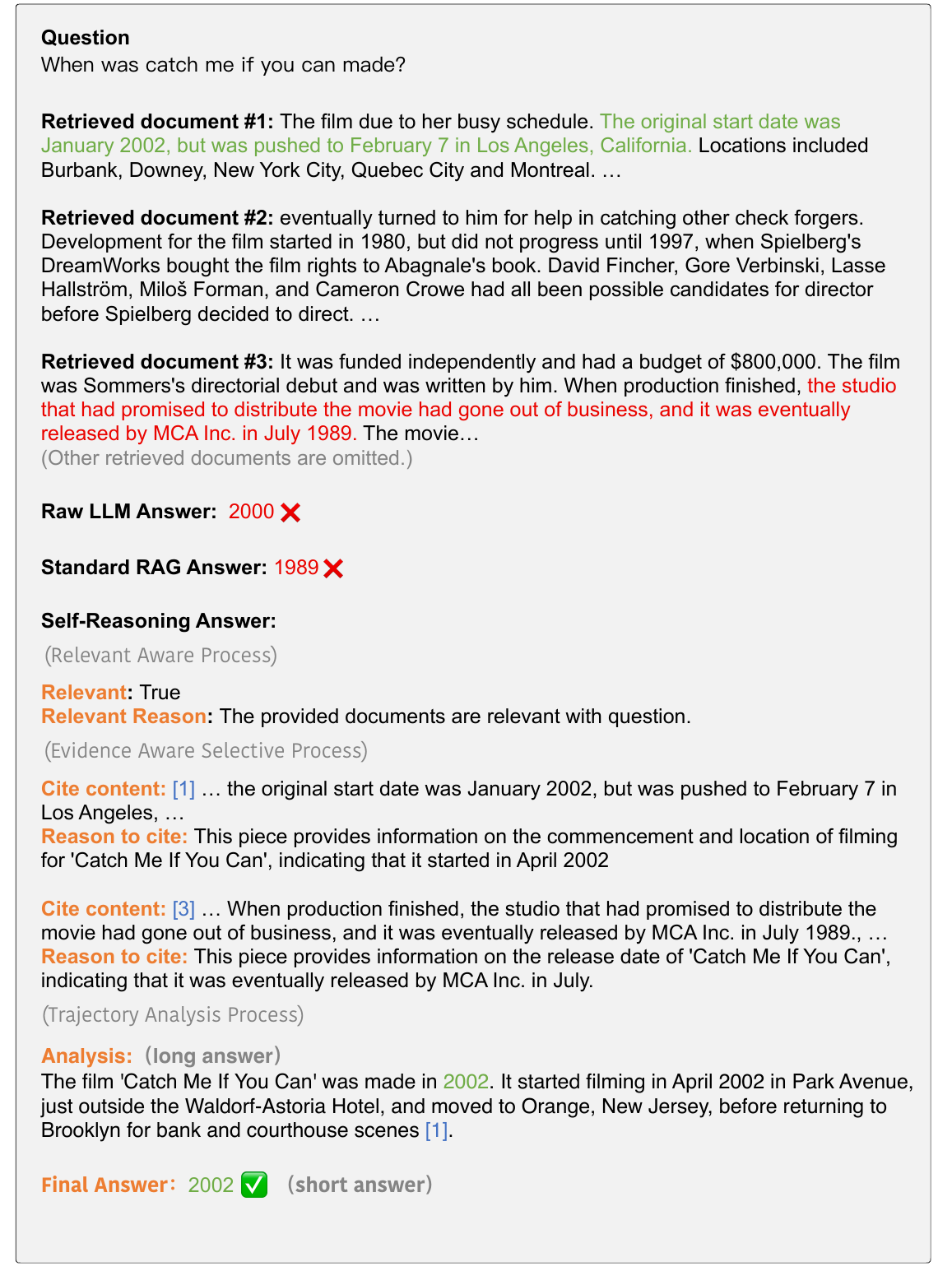}
    \caption{A Case Study. We present the self-reasoning trajectories generated by our framework during inference and demonstrate how they can logically generate the correct answer.}
    \label{fig:case study}
\end{figure*}

{\label{app:case study}}
In our case study, as illustrated in Figure \ref{fig:case study}, we compare the responses generated by the raw LLM, the standard RALM (e.g. LLaMA2 with retrievals), and our \textsc{self-reasoning} method. The challenge involves reconciling information from multiple retrieved documents to provide a correct answer as the retrieved documents contained noisy data.

The response from the raw LLM (e.g., LLaMA2) suggested that the film was made in 2000, based on its inherited knowledge. However, this answer is incorrect and is a hallucination generated by the LLM. The standard RALM approach yielded 1989 as the production date. This answer was based on unrelated details from the retrieved documents, showing a lack of context-specific understanding and robustness for noisy retrieved documents. 

Our \textsc{self-reasoning} framework provided a comprehensive approach by assessing the relevance and context of retrieved documents. First, in the relevance-aware process, the documents were identified as relevant based on their content regarding the production dates and events surrounding the film. Second, in the evidence-aware selective process, the model retrieved the first documents, which highlighted the original start date as January 2002, with filming commencing in February 2002 (highlighted in green in the figure). This information was crucial in establishing the timeline for the film's production. The model can also understand of the difference between the production date and the release date in the third retrieved document (highlighted in red in the figure). In the trajectory analysis process, the correct timeline was deduced by piecing together self-generated trajectories, leading to the conclusion that the film \textit{Catch Me If You Can} was indeed produced in 2002. As the case illustrated above, by leveraging relevant documents and focusing on contextual evidence, our \textsc{self-reasoning} framework can achieve a precise and well-supported answer, highlighting its utility and robustness in complex information retrieval tasks.

\subsection{More Analysis on Ablation Study }\label{app:ablation}
\subsubsection*{Effectiveness of Gradual Learning.}
Further, we validate the effect of gradual learning. Rather than training an LLM with a stage-by-stage approach, we initially concatenate the reasoning trajectories from all stages and put them into the LLM for end-to-end training. As shown in Table \ref{tab:ablation}, the performance decline can be observed in three datasets, suggesting that gradual learning can help improve performance.

\subsubsection*{Effectiveness of Quality Control.}
The effect of quality control on data generation is also evaluated in our work. Instead of using the filtered high-quality training samples, we randomly sampled 2,000 unfiltered training samples generated by GPT-4. As shown in Table \ref{tab:ablation}, the substitution of unfiltered training data leads to the degradation of the model results.

\subsection{Human Evaluation}{\label{app:human}}
We randomly sample 100 examples from the ASQA dataset and annotate the outputs of selected models. Each sample is then assigned to two people for annotation. Each annotator is required to verify the citation recall and citation precision acorrding to our provided scheme. The annotation scheme is inspired by \cite{gao-etal-2023-enabling} as follows:

\paragraph{Citation Recall.}
The annotators are shown the question $q$, the statement $s_i$, and all of its citations $C_i$, and they assess if the set of citations fully support the statement (recall=1) or if they do not support all the claims (recall=0). We calculate the overall recall score for the model by averaging the recall scores of all statements.

\paragraph{Citation Precision.}
The annotators are shown the question $q$ and a statement $s_i$ and one of its citation ${c^{(k)}_i} \in C_i $. We ask the annotator if the citation \textit{fully supports}, \textit{partially supports}, or \textit{does not support} the generated claims in $s_i$. Citation ${c^{(k)}_i}$ has a citation precision of 1 if $s_i$ has a recall of 1, and ${c^{(k)}_i}$ fully or partially supports $s_i$. Finally, we calculate the overall precision score for the model by averaging the precision scores of all statements.

\subsection{The Pseudo-code of Data Quality Control}
\xia{The details and pseudo-code of data quality control are illustrated in Algorithm \ref{data:qc}. In the table, $\mathcal{D}_0$ is the origin self-reasoning dataset generated by GPT-4. $P$ is the program to evaluate the answer according to the metrics. $T$ is the off-the-shelf tool \cite{gao-etal-2023-enabling} to calculate the citation scores. $N$ is the number of total samples from the origin dataset, and $d_i$ is the $i$-th training sample. $s_r$ and $s_c$ are citation recall and precision scores for each sample respectively. $\delta_{r}$ and $\delta_{p}$ are thresholds of citation recall and precision respectively. $\mathcal{D}_l$ is filtered trajectory data for long-form QA, and $\mathcal{D}_s$ is filtered dataset for short-form QA and fact verification dataset.  $\mathcal{D}_{sr}$ is the final high quality self-reasoning training dataset.}

\subsection{Latency Analysis}
We conduct extra experiments to measure inference latency across our approach, Self-RAG, and GPT-4. The results demonstrated that our method achieved better performance while maintaining latency comparable to Self-RAG. The results of average inference latency per question for the NQ and ASQA datasets are shown in Table \ref{tab:latency}. 

\begin{table}[t]
\center
  \begin{tabularx}{0.42\textwidth}{p{2.8cm}<{\centering}p{1.6cm}<{\centering}p{1.6cm}<{\centering}}
    \toprule
    { \textbf{Models} } &  \textbf{NQ} &  \textbf{ASQA}   \\
    \midrule
LLaMA2  & 0.19 & 1.92  \\

Self-RAG  & 2.08 & 2.38  \\
\textsc{Self-Reasoning}  & 2.19 &2.32  \\
\hdashline
GPT-4  & 35.5 & 40.5  \\

  \bottomrule
\end{tabularx}
\caption{The latency experiment results on NQ and ASQA datasets. The average inference latency per question is evaluated using 7B-parameter models.}
  \label{tab:latency}
\end{table}

\end{document}